\documentclass[11pt]{article}
\usepackage{coling2018}
\usepackage{times}
\usepackage{url}
\usepackage{latexsym}

\usepackage{tabularx}
\usepackage{float}
\usepackage{xspace}
\usepackage{graphicx}
\usepackage{bmpsize}
\graphicspath{ {images/} }

\newcommand{\App}{SetExpander\xspace}

\title{Term Set Expansion based on Multi-Context Term Embeddings:\\
an End-to-end  Workflow  }
\author{Jonathan Mamou,$^1$ Oren Pereg,$^1$ Moshe Wasserblat,$^1$ Ido Dagan,$^2$ Yoav Goldberg,$^2$ \\
    \textbf{Alon Eirew,$^1$ Yael Green,$^1$ Shira Guskin,$^1$ Peter Izsak,$^1$ Daniel Korat$^1$} \\
  $^1$Intel AI Lab, Israel \\
  $^2$Department of Computer Science, Bar-Ilan University,  Ramat Gan, Israel \\
  $^1${\tt firstname.lastname@intel.com} \\
  $^2${\tt \{dagan,yogo\}@cs.biu.ac.il}\\
}

\date{}

\begin{document}

\maketitle

\begin{abstract}
\blfootnote{
    \hspace{-0.65cm}  
    This work is licensed under a Creative Commons 
    Attribution 4.0 International License.
    License details:
    \url{http://creativecommons.org/licenses/by/4.0/}.
}
We present \App, a corpus-based system for expanding a seed set of terms into a more complete set of terms that belong to the same semantic class. 
\App implements an iterative end-to end workflow for term set expansion. It enables users to easily select a seed set of terms, expand it, view the expanded set, validate it, re-expand the validated set and store it, thus simplifying the extraction of domain-specific fine-grained semantic classes. \App has been used for solving real-life use cases including integration in an automated recruitment system and an issues and defects resolution system.\footnote{A video demo of \App is available at \url{https://drive.google.com/open?id=1e545bB87Autsch36DjnJHmq3HWfSd1Rv}  (some images were blurred for privacy reasons).}
\end{abstract}

\section{Introduction}
\label{sec:introduction}

Term set expansion is the task of expanding a given partial set of terms into a more complete set of terms that belong to the same semantic class. For example, given the partial set of personal assistant application terms like `Siri' and `Cortana' as seed, the expanded set is expected to include additional personal assistant application terms such as `Amazon Echo' and `Google Now'.
Many NLP-based information extraction applications, such as relation extraction or document matching, require the extraction of terms belonging to fine-grained semantic classes as a basic building block. 
A practical approach to extracting such terms 
is to apply a term set expansion system. The input seed set for such systems may contain as few as two to ten terms which is practical to obtain.

\App uses a corpus-based approach based on the {\it distributional similarity hypothesis}~\cite{harris1954distributional}, stating that semantically similar words appear in similar contexts. Linear bag-of-words context is widely used to compute semantic similarity.
However, it typically captures more {\it topical} and less {\it functional} similarity, while for the purpose of set expansion, we need to capture more functional and less topical similarity.\footnote{We use the terminology introduced by~\cite{turney2012domain}: the {\it topic} of a term is characterized by the nouns that occur in its neighborhood while the {\it function} of a term is characterized by the syntactic context that relates it to the verbs that occur in its neighborhood.} For example, given a seed term like the programming language 'Python', we would like the expanded set to include other programming languages with similar characteristics, but we would not like it to include terms like `bytecode' or `high-level programming language' despite these terms being semantically related to `Python' in linear bag-of-words contexts.

Moreover, for the purpose of set expansion, a seed set contains more than one term and the terms of the expanded set are expected to be as functionally similar to {\it all} the terms of the seed set as possible.
For example, `orange' is functionally similar to `red' (color) and to `apple' (fruit), but if the seed set contains both `orange' and `yellow' then only `red' should be part of the expanded set. However, we do not want to capture only the term sense; we also wish to capture the granularity within a category. For example, `orange' is functionally similar to both `apple' and `lemon'; however, if the seed set contains `orange' and `banana' (fruits), the expanded set is expected to contain both `apple' and `lemon'; but if the seed set is `orange' and `grapefruit' (citrus fruits), then the expanded set is expected to contain `lemon' but not `apple'.

While term set expansion has received attention from both industry and academia, there are only a handful of available implementations. Google Sets was one of the earliest applications for term set expansion. It used methods like latent semantic indexing to pre-compute lists of similar words (now discontinued). Word Grab Bag\footnote{\url{www.wordgrabbag.com}} builds lists dynamically using word2vec embedding’s based on bag-of-word contexts, but its algorithm is not publicly described. State-of-the-art research techniques are based on computing semantic similarity between seed terms and candidate terms in a given corpus and then constructing the expanded set from the most similar terms~\cite{sarmento2007more,shen2017setexpan}. 

Relative to prior work, the contribution of this paper is twofold. First, it describes an iterative end-to-end workflow that enables users to select an input corpus, train multiple embedding models and combine them; after which the user can easily select a seed set of terms, expand it, view the expanded set, validate it, iteratively re-expand the validated set and store it.
Second, it describes the \App system which is based on a novel corpus-based set expansion algorithm developed in-house; this algorithm combines multi-context term embeddings to capture different aspects of semantic similarity and to make the system more robust across different domains. The SetExpander algorithm is briefly described in Section~\ref{sec:algo}. 
Our system has been used for solving several real-life use cases. One of them is an automated recruitment system that matches job descriptions with job-applicant resumes. 
Another use case involves enhancing a software development process by detecting and reducing the amount of duplicated defects in a validation system. Section~\ref{sec:usecases} includes a detailed description of both use cases.
The system is distributed as open source software under the Apache license as part of NLP Architect by Intel AI Lab~\footnote{\url{http://nlp_architect.nervanasys.com/term_set_expansion.html}}

\section{Term Set Expansion Algorithm Overview}
\label{sec:algo}

Our approach is based on representing any term of a training corpus using word embeddings in order to estimate the similarity between the seed terms and any candidate term. 

Noun phrases provide good approximation for candidate terms and are extracted in our system using a noun phrase chunker.\footnote{\url{http://nlp_architect.nervanasys.com/chunker.html}} Term variations, such as aliases, acronyms and synonyms, which refer to the same entity, are grouped together. We use a heuristic algorithm that is based on text normalization, abbreviation web resources, edit distance and word2vec similarity. For example, {\it New York, New-York, NY, New York City} and {\it NYC} are grouped together forming a single term group. Then, we use term groups as input units for embedding training; it enables obtaining more contextual information compared to using individual terms, thus enhancing the robustness of the embedding model. 

Our basic algorithm version follows the standard unsupervised set expansion scheme. Terms are represented by their linear bag-of-words window context embeddings using the word2vec toolkit.\footnote{\url{http://code.google.com/archive/p/word2vec}} At expansion time, given a seed of terms, the most similar terms are returned where similarity is estimated by the cosine similarity between the centroid of the seed terms and each candidate term.
While word2vec typically uses a linear bag-of-words window context around the focus word, the literature describes other possible context types (Table~\ref{table:context:example}). We found that indeed in different domains, better similarities are found using different context types. The different contexts thus complement each other by capturing different types of semantic relations.
Typically, explicit list contexts work well for the automated recruitment system use case, while unary patterns contexts work well for the issues and defects resolution use case (Section~\ref{sec:usecases}).
To make the system more robust, we extended the basic algorithm to combine multi-context embeddings. Terms are represented with arbitrary context embeddings trained using the generic word2vecf toolkit.\footnote{\url{http://bitbucket.org/yoavgo/word2vecf}} Taking the similarity scores between the seed terms and the candidate terms according to each of the different contexts as features, a Multilayer Perceptron (MLP) binary classifier predicts whether a candidate term should be part of the expanded set, where training and development term lists are used for the MLP training. 
The MLP classifier is implemented on top of Neon,\footnote{\url{http://github.com/NervanaSystems/neon}} the Intel Nervana Deep Learning Framework. The performance of the algorithm was first evaluated by MAP@$n$ (Mean Average Precision at $n$). MAP@10, MAP@20 and MAP@50 on an English Wikipedia dataset~\footnote{Dataset is described at \url{http://nlp_architect.nervanasys.com/term_set_expansion.html}.} are respectively 0.83, 0.74 and 0.63.

\begin{table}[h!]

\begin{tabularx}{\textwidth}{|X|X|l|X|}
    \hline
    \bf{Context Type} &  \bf{Example sentence} & \bf{Focus term} & \bf{Context units} \\ \hline
    Linear bag-of-words~\cite{mikolov2013distributed} & {\it Siri uses voice queries and a  natural language user interface.} & {\it Siri} & {\it uses, voice queries, natural language user interface}\\ \hline
    Explicit lists~\cite{sarmento2007more} & {\it Experience in Image processing, Signal processing, Computer Vision} & {\it Image processing} & {\it Signal processing, Computer Vision} \\ \hline
    Syntactic dependency~\cite{levy2014dependency} & {\it Turing studied as an undergraduate ... at King's College, Cambridge.} & {\it studied} & {\it (Turing/nsubj), (undergraduate\-/\-prep\_as), (King's College/prep\_at)} \\ \hline
    Symmetric patterns~\cite{schwartz2015symmetric} & {\it Apple and Orange juice drink} & {\it Apple} & {\it Orange} \\ \hline
    Unary patterns~\cite{rong2016egoset} & {\it In the U.S. state of Alaska ...} & {\it Alaska} & {\it U.S. state of \_\_} \\ \hline
\end{tabularx}
\caption{Examples of extracted contexts per context type.}
\label{table:context:example}
\end{table}

\section{System Workflow and Application}
\label{sec:workflow}

\begin{figure}[ht]
\includegraphics[width=\textwidth]{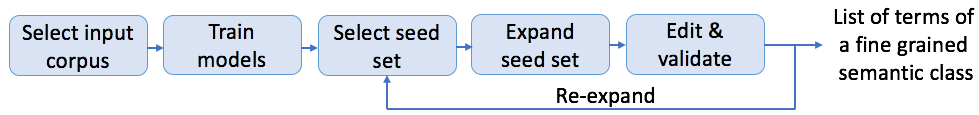}
\centering
\caption{\App end-to-end workflow.}
\label{flow_fig}
\end{figure}

\begin{figure}[ht]
\includegraphics[width=\textwidth]{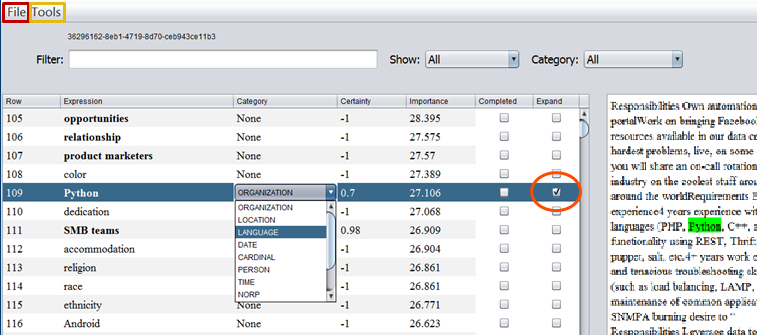}
\centering
\caption{\App user interface for seed selection and expansion.}
\label{app1_fig}
\end{figure}

\begin{figure}[ht]
\includegraphics[width=0.25\textwidth]{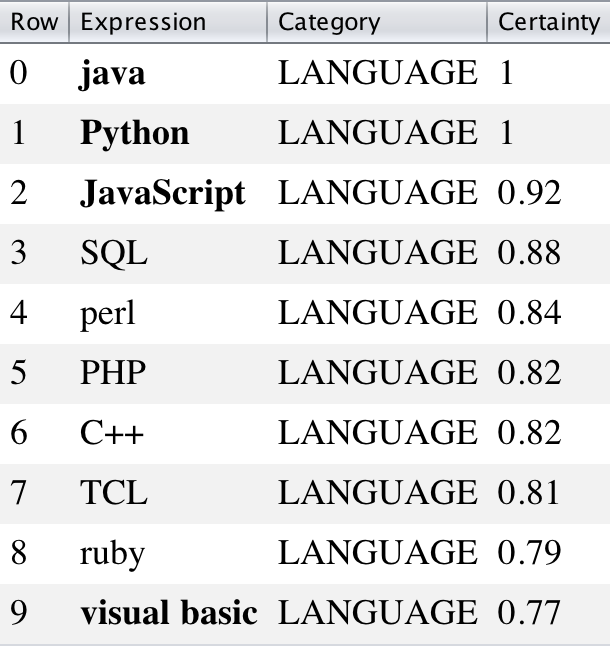}
\centering
\caption{\App user interface for expansion results output. Seed terms are `java' and `python'.
}
\label{app2_fig}
\end{figure}

This section describes the iterative end-to-end workflow of \App as depicted in Figure~\ref{flow_fig}. Each step of the flow is performed by the user using the system's user interface (Figures~\ref{app1_fig} and~\ref{app2_fig}). 
The first two steps of the flow are to \textbf{select an input corpus} and to \textbf{train models}. The ``train models'' step extracts term groups from the corpus and trains the combined term groups embedding models (Section \ref{sec:algo}). 
Next, the user is able to \textbf{select a seed set} for expansion. Figure~\ref{app1_fig} shows the seed set selection and expansion user interface. Each row in the displayed table corresponds to a different term group. The term group names are displayed under the `Expression' column.
The `Filter' text box is used for searching for specific term groups. 
Upon selecting (clicking) a term group, the context view on the right hand side displays text snippets from the input corpus that include terms that are part of the selected term group (highlighted in green in Figure~\ref{app1_fig}). 
The user can create a seed set assembled from specific term groups by checking their `Expand' checkbox (see the red circle in Figure~\ref{app1_fig}). The user can select or set a name for the semantic category of the seed set (see drop down list in Figure~\ref{app1_fig}). Once the seed set is assembled, the user can \textbf{expand the seed set} by selecting the Expand option in the tools menu (not shown). 

Figure~\ref{app2_fig} shows the output of the expansion process. The expanded term groups are highlighted in green. The Certainty score represents the relatedness of each expanded term group to the seed set. This score is determined by the MLP classifier (Section \ref{sec:algo}). The Certainty scores of term groups that were manually selected as part of the seed set, are set to 1. 
The user can perform \textbf{re-expansion} by creating a new seed set based on the expanded terms and the original seed set terms. The user is also able to validate each expanded item by checking the ``Completed'' checkbox. The validated list can then be \textbf{saved} and later used as a fine-grained taxonomy input to external information-extraction systems. 

\section{Field Use Cases}\label{sec:usecases}

This section describes two use cases in which \App has been successfully used. 

\subsection{Automated Recruitment System} 
Human matching of applicant resumes to open positions in organizations is time-consuming and costly. Automated recruitment systems enable recruiters to speed up and refine this process. The recruiter provides an open position description and then the system scans the organization’s resume repository searching for the best matches. One of the main features that affect the matching is the skills list, for example, a good match between an applicant and an open position regarding specific programming skills or experience using specific tools is significant for the overall matching. However, manual generation and maintenance of comprehensive and updated skills lists is tedious and difficult to scale. \App was integrated into such a recruitment system. Recruiters used the system's user interface (Figures~\ref{app1_fig} \& \ref{app2_fig})  to generate fine-grained skills lists based on small seed sets for eighteen engineering job position categories. We evaluated the recruitment system use case for different skill classes. The system achieved a precision of 94.5\%, 98.0\% and 70.5\% at the top 100 applicants, for the job position categories of Software Machine Learning Engineer, Firmware Engineer and ADAS Senior Software Engineer, respectively.

\subsection{Issues and Defects Resolution}  
Quick identification of duplicate defects is critical for efficient software development. The aim of automated issues and defects resolution systems is to find duplicates in large repositories of millions of software defects used by dozens of development teams. This task is challenging because the same defect may have different title names and different textual descriptions. The legacy solution relied on manually constructed lists of tens of thousands of terms, which were built over several weeks. Our term set expansion application was integrated into such a system and was used for generating domain specific semantic categories such as product names, process names, technical terms, etc. The integrated system enhanced the duplicate defects detection precision by more than 10\% and sped-up the term list generation process from several weeks to hours.
\section{Conclusion}
\label{sec:conclusion}
We presented \App, a corpus-based system for set expansion which enables users to select a seed set of terms, expand it, validate it, re-expand the validated set and store it. The expanded sets can then be used as a domain specific semantic classes for downstream applications. Our system was used in several real-world use cases, among them, an automated recruitment system and an issues and defects resolution system.

\section*{Acknowledgements}
This work was supported in part by an Intel ICRI-CI grant. The authors are grateful to Sapir Tsabari from Intel AI Lab for her help in the dataset preparation. 

\bibliography{references}
\bibliographystyle{acl}

\end{document}